\documentclass[review]{elsarticle}
\pdfoutput=1

\usepackage{geometry}
\usepackage{diagbox}
\usepackage{subfigure}
\usepackage{amssymb}
\usepackage{amsmath}
\usepackage{graphicx}
\usepackage[dvipsnames]{xcolor}
\usepackage{lineno}
\usepackage[breaklinks]{hyperref}
\usepackage{microtype}
\usepackage{multirow}
\usepackage{gensymb}

\begin{document}

\begin{frontmatter}

\title{Animal Behavior Classification via Accelerometry Data and Recurrent Neural Networks}

\author[af]{Liang Wang}
\author[data61]{Reza Arablouei\corref{cor1}}
\author[dpi]{Flavio A. P. Alvarenga}
\author[af]{Greg J. Bishop-Hurley}
\cortext[cor1]{Corresponding author}
\address[af]{Agriculture and Food, CSIRO, St Lucia QLD 4067, Australia}
\address[data61]{Data61, CSIRO, Pullenvale QLD 4069, Australia}
\address[dpi]{NSW Department of Primary Industries, Armidale NSW 2350, Australia}

\begin{abstract}

We study the classification of animal behavior using accelerometry data through various recurrent neural network (RNN) models. We evaluate the classification performance and complexity of the considered models, which feature long short-time memory (LSTM) or gated recurrent unit (GRU) architectures with varying depths and widths, using four datasets acquired from cattle via collar or ear tags. We also include two state-of-the-art convolutional neural network (CNN)-based time-series classification models in the evaluations. The results show that the RNN-based models can achieve similar or higher classification accuracy compared with the CNN-based models while having less computational and memory requirements. We also observe that the models with GRU architecture generally outperform the ones with LSTM architecture in terms of classification accuracy despite being less complex. A single-layer uni-directional GRU model with 64 hidden units appears to offer a good balance between accuracy and complexity making it suitable for implementation on edge/embedded devices.

\end{abstract}


\end{frontmatter}

\section{Introduction}

Monitoring the behavior of livestock is crucial for effective management. Behavior information can provide insights into the productivity, health, and welfare of animals. However, monitoring animal behavior manually is time-consuming, labor-intensive, error-prone, and potentially disturbing to animals. Therefore, using wearable sensors to automatically classify animal behavior is of practical importance. Most animal behaviors are recognizable through body movements that can be sensed using accelerometers.

The early approaches using accelerometry data to classify animal behavior used hand-tuned thresholds in conjunction with some standard statistical features in the time or frequency domains. In \cite{williams2019application}, the drinking behavior is identified using the mean of the accelerometer readings in the $z$ (front-to-back) axis and the variance of the accelerometer readings in the $x$ (vertical) axis. The results show that the proposed method can distinguish the drinking behavior from the other behaviors with a relatively high accuracy. \citet{busch2017determination} extract a few standard statistical features from triaxial accelerometry data collected by collar tags on dairy cattle and calculate the relevant thresholds to distinguish lying from standing with good accuracy. In this work, the threshold values are computed manually through analyzing the associated data.

Accelerometer readings are time-series data as the order of the sensed values matters. Several time-series classification algorithms have been proposed in recent years~\citep{bagnall2017great}. The most popular ones are the nearest neighbor methods coupled with parametric distance functions \citep{lines2015time} and the ensemble methods \citep{bagnall2015time,bostrom2015binary,schafer2015boss,kate2016using} including random forest \citep{baydogan2013bag,deng2013time}. \citet{rahman2018cattle} use several statistical features in time and frequency domains to train a random forest classifier that can differentiate three cattle behaviors of grazing, standing, and ruminating. \citet{smith2016behavior} implement a set of binary classifiers, i.e., support vector machine (SVM), naive Bayes, $k$-nearest neighbors, logistic regression, and random forest, for animal behavior classification. They use standard statistical features in time and frequency domains together with a few information-theoretic features. The results show that the ensembles of binary classifiers may offer improvement in classification accuracy over the corresponding multi-class classifiers. \citet{arablouei2021situ} devise a small set of computationally-efficient and informative features based on the accelerometry data's statistical and spectral properties. The validations show that resource-efficient discriminative models can yield good classification accuracy with the proposed features.

The above-mentioned machine-learning-based methods for animal behavior classification require careful feature engineering to achieve good results. The features used in \cite{rahman2018cattle} and \cite{smith2016behavior} stem from the domain knowledge and intuition, hence do not necessarily represent the internal structure of the data. In addition, such methods require the extraction of the features to be accomplished before training any classification model. As a result, the feature extraction and classification stages are disjointed.

End-to-end deep neural networks (DNNs) have demonstrated good performance in classification tasks with complex input data, e.g., image classification \cite{lecun2015deep}, where feature engineering is inherently difficult. Therefore, substantial research has been conducted on using various DNN models for time-series classification \cite{wang2017time,fawaz2019deep}.

\citet{rahman2016comparison} use an auto-encoder to learn useful feature representations of accelerometer time-series data in an unsupervised fashion. An SVM classifier trained with the learned features is shown to outperform the one trained with the hand-crafted features. This work achieves data-driven feature engineering while the feature extraction and the classification are still separate steps.

\citet{kasfi2016convolutional} introduce an end-to-end cattle behavior classifier using a convolutional neural network (CNN) model. In this model, the parameters of the feature extraction and classification modules are estimated simultaneously. The proposed algorithm's accuracy is similar to that of the one proposed in \cite{rahman2016comparison}, albeit it is more efficient. \citet{peng2019classification} use a long short-time memory (LSTM) recurrent neural network (RNN) model to classify cattle behavior. The model is trained to classify cattle behavior considering three different time window sizes of $3.2$s, $6.4$s, and $12.8$s. The results show that the utilized LSTM classifier is superior to a simple CNN classifier and the best performance is achieved when the window size is $3.2$s. Following up this work, \citet{peng2020dam} use a similar LSTM model to detect calving-related behavior using accelerometry data. Unfortunately, the authors of \cite{peng2019classification} and \cite{peng2020dam} do not provide any further detail regarding the specifications of their models.

In this paper, we examine the performance of RNN models in classifying animal behavior using accelerometry data. We explore different RNN model architectures, i.e., LSTM and gated recurrent unit (GRU) with various numbers of layers and units, to characterize their performance as well as the underlying trade-offs between model complexity and accuracy. Understanding the interplay between the model complexity and classification accuracy is critical when the behavior classification is to be executed on devices with limited computational, memory, or energy resources. By characterizing the accuracy-complexity trade-off for RNN-based classifiers, we assess the feasibility of their implementation on edge or embedded devices and identify the most suitable models, which attain the best balance between accuracy and complexity.

The remainder of the paper is organized as follows. In Section~\ref{sec:data}, we explain the datasets used in this paper alongside how they have been collected. In Section~\ref{sec:model}, we describe the classification model architecture, the RNN variants, and the related hyper-parameters that we consider in this work. In Sections~\ref{sec:eva} and \ref{sec:discuss}, we evaluate the performance of the proposed RNN-based classifiers in comparison with two popular CNN-based classifiers and discuss the associated accuracy-complexity trade-offs. We draw conclusions in Section~\ref{sec:conclusion}.

\section{Datasets} \label{sec:data}

In this section, we describe the specifications of the datasets that we utilize for evaluating our proposed animal behavior classification models as well as the processes of their generation.

We have conducted several data collection experiments with cattle wearing collar or ear tags. The first experiment took place in August 2018 at the Commonwealth Scientific and Industrial Research Organisation (CSIRO) FD McMaster Laboratory Chiswick, near Armidale, NSW, Australia (30\degree36'28.17"S, 151\degree32'39.12"E). There is a detailed description of the experiment and the equipment used including the collar tag in \cite{arablouei2021situ}. The accelerometry data was collected from ten steers wearing collar tags. We refer to the associated dataset as Arm18. Another experiment was conducted in March 2020 at the same facility while the accelerometry data was recorded from seven cattle using both collar and ear tags. We refer to the associated datasets as Arm20c and Arm20e where the former is collected by collar tags and the latter by ear tags. A similar experiment was run at CSIRO's Lansdown Research Station, Woodstock, Queensland, Australia (19\degree39'26.41"S, 146\degree50'5.88"E) in September 2019 with nineteen cattle wearing collar tags. The animals in this experiments were heifers and steers of mainly Brangus and Droughtmaster breeds. We refer to the associated dataset as Lan19.

\begin{figure}
    \centering
    \includegraphics[width=14cm]{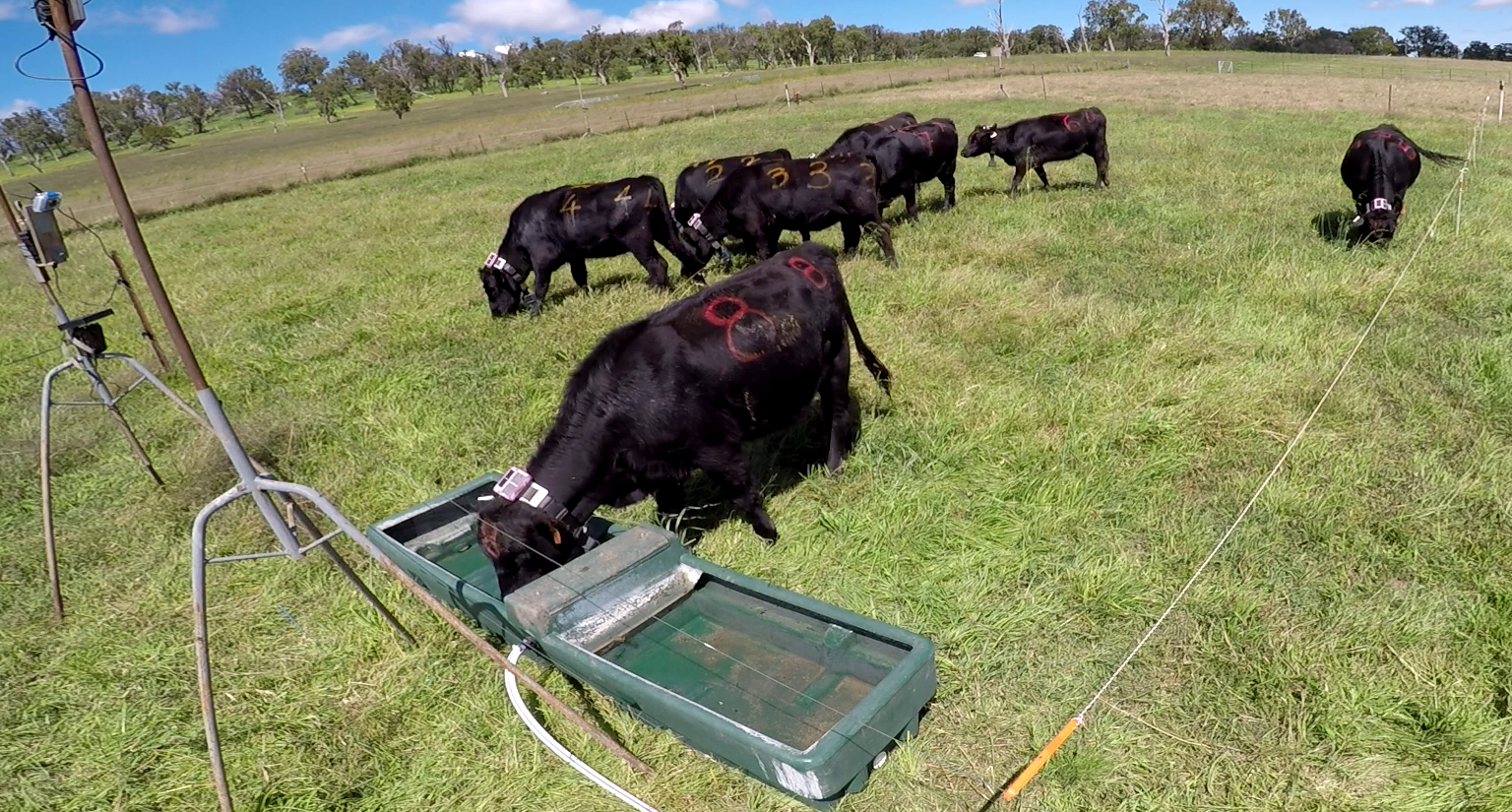}
    \caption{Cattle on paddock during the experiment leading to the Arm20c and Arm20e datasets.}
    \label{area}
\end{figure}

\begin{figure}
    \centering
    \subfigure[collar tag]{\includegraphics[width=6.55cm]{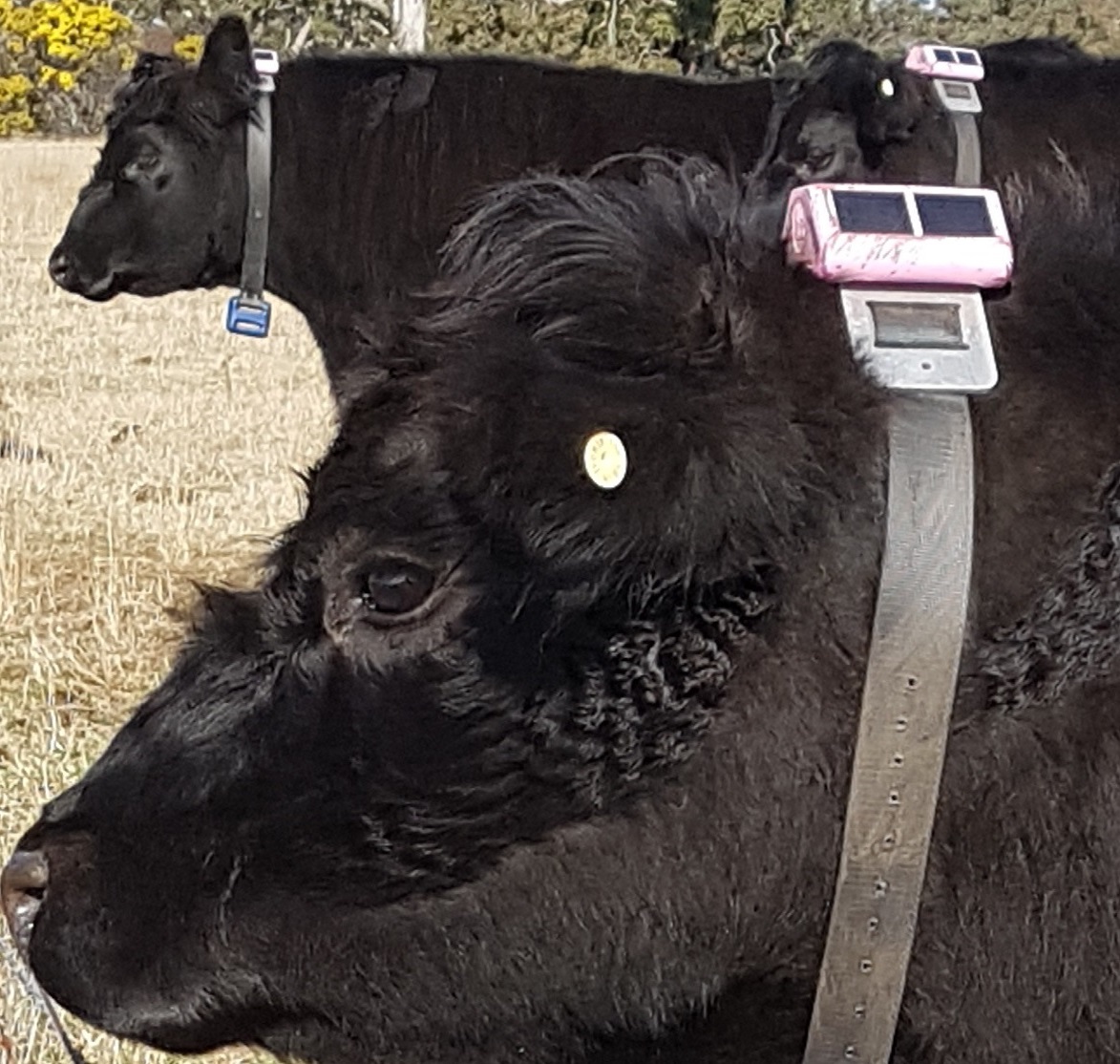}\label{collar}}
    \qquad
    \subfigure[ear tag]{\includegraphics[width=6.55cm]{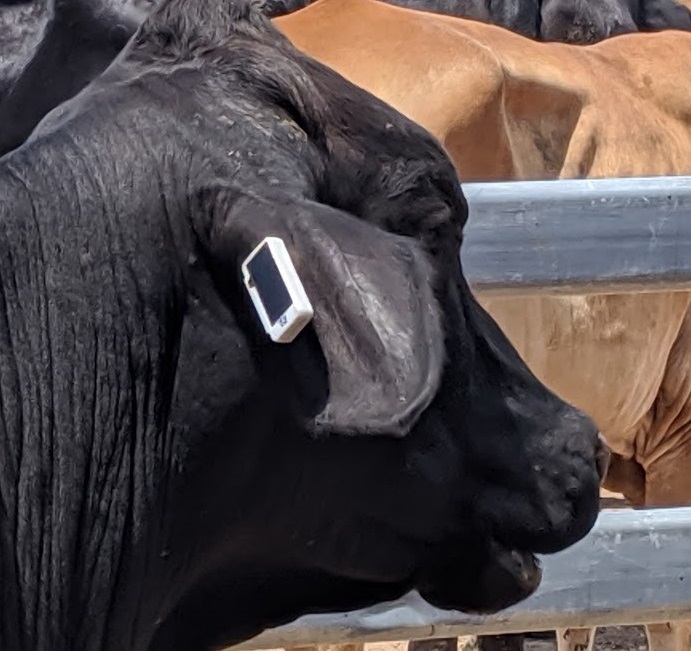}\label{eartag}}
    \caption{Angus heifers wearing collar (b) and ear (c) tags used for data collection.}
    \label{tags}
\end{figure}

Fig.~\ref{area} shows the paddock and the cattle used for the experiment that produced the Arm20c and Arm20e datasets. Figs.~\ref{collar} and~\ref{eartag} show cattle wearing the collar and ear tags. The experiments were approved by the CSIRO FD McMaster Laboratory Chiswick Animal Ethics Committee with the animal research authority number 17/20. 

The smart ear tag used for data collection, specifically, the Arm20e dataset, is a purpose-built sensor node jointly developed by CSIRO and a commercial partner to serve both research and industry applications. It houses a wealth of sensors and communication capabilities. Its main components are microcontroller, triaxial accelerometer, satellite communication interface, on-board memory, solar panel, and battery. The microcontroller is Nordic nRF52840 system-on-chip with a 64MHz ARM Cortex-M4F processor, 1MB of flash memory, 256kB of RAM, a floating-point unit, and Bluetooth 5 interface. The accelerometer is a Bosch Sensortec BMA400 micro electromechanical system (MEMS) device with an ultra-low power consumption (max $14\mu$A). It measures the instantaneous acceleration in three orthogonal spatial axes. On each axis, it senses minuscule changes in the capacitance between a fixed electrode and a proof mass that is displaced by any force applied to the device's supporting frame.

\begin{figure}
    \centering
    \subfigure[PCB]{\includegraphics[width=5.1cm]{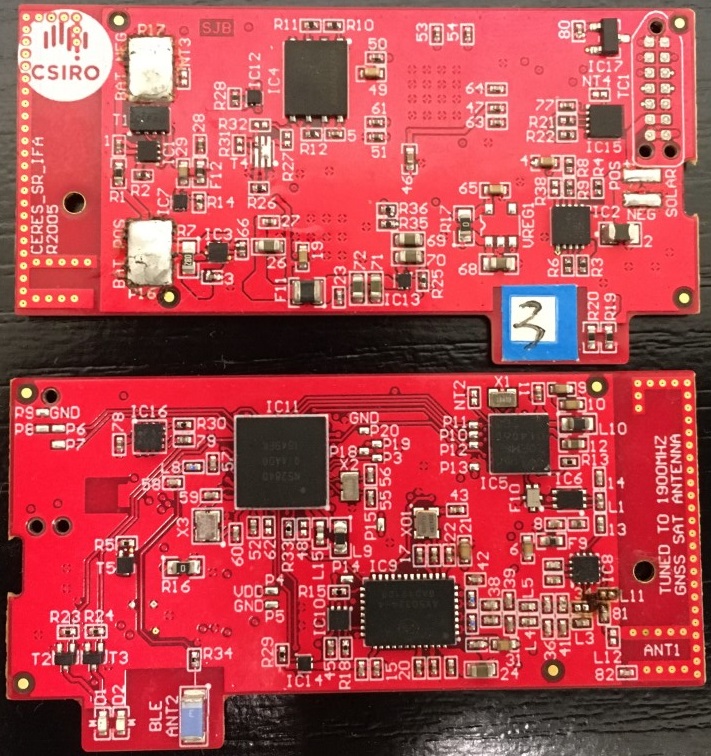}\label{pcb}}
    \qquad
    \subfigure[block diagram]{\includegraphics[width=8cm]{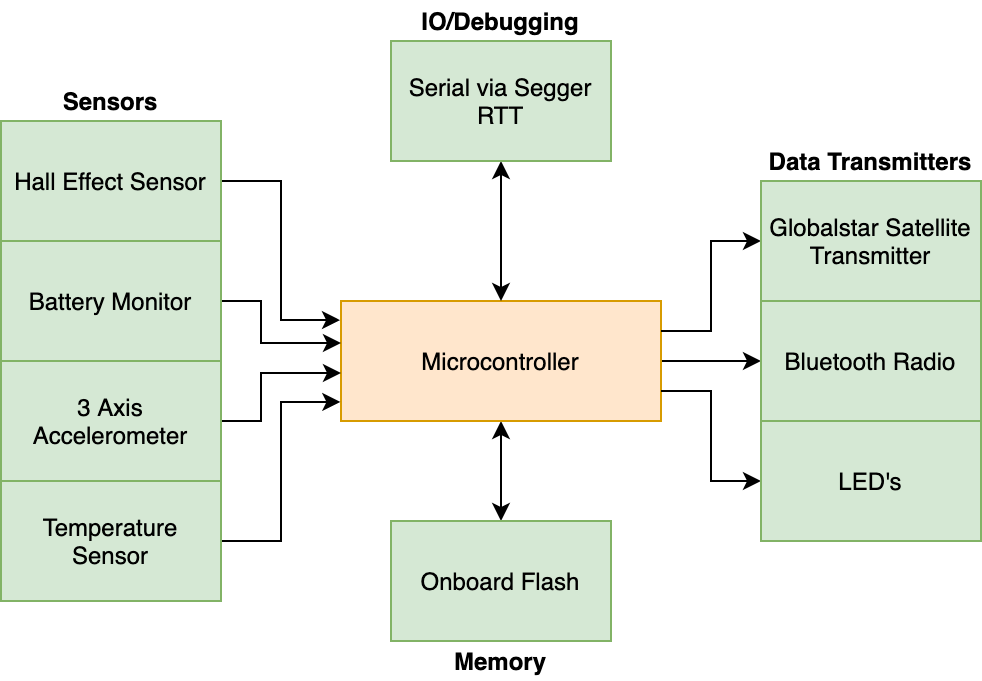}\label{blkdgrm}}
    \caption{The PCB of the ear tag used for data collection (a) and the block diagram of its main components (b).}
    \label{ceres}
\end{figure}

The ear tag is powered by a $3.2$V, $170$mAh battery pack that is recharged using a solar panel installed on the tag's case. The operating system running on the microcontroller is Zephyr~\cite{zephyr}. As the on-board memory of the ear tag is limited, to log the accelerometery data, we stream the data via a generic attribute profile (GATT) Bluetooth link to the collar tag installed on the same animal. Fig.~\ref{pcb} contains the pictures of both sides of the ear tag's printed circuit board (PCB) and Fig.~\ref{blkdgrm} depicts the block diagram of the ear tag's main components.

In all data collection experiments, the tags recorded triaxial accelerometer readings. The accelerometer sampling rate was $50$ Hz for the collar tags and $62.5$ Hz for the ear tags. We manually annotated sections of the accelerometry data by observing the behavior of the cattle. The behaviors are mutually exclusive. In the Arm18 dataset, there are four behavior classes of interest, i.e., grazing, ruminating, resting, and other. In the Lan19 dataset, four behavior classes of interest are grazing, walking, ruminating/resting, and other. In the Arm20c and Arm20e datasets, there are three behavior classes of interest, i.e., grazing, ruminating/resting, and other. In the Arm18, Arm20c, and Arm20e datasets, the \emph{other} behavior class is the collection of all behaviors other than grazing, resting, and ruminating. In the Lan19 dataset, the \emph{other} behavior class represents all behaviors except grazing, walking, resting, and ruminating.

We divide The annotated accelerometry data into non-overlapping segments each containing $256$ consecutive triaxial readings. The segment width of $256$ corresponds to about $5.12$s in the collar tag data and about $4.1$s in the ear tag data. Therefore, the accelerometer readings of each dataset can be arranged into a three-way tensor of dimensions $(N, 256, 3)$ where $N$ is the total number of segments. Table~\ref{tab:dataset} shows the number of segments (data samples) for each behavior class in each dataset.

\begin{table}[htbp] 
\caption{The sample number of each behavior class in the considered datasets} 
\label{tab:dataset}
\centering
\begin{tabular}{l c c c c c}
\hline
& \multicolumn{4}{c}{dataset} &\\
behavior   & Arm18 & Lan19 & Arm20c & Arm20e\\
\hline
grazing    & 7109  & 70402 & 6156   & 6047\\
walking    & -     & 16700 & -      & -\\
ruminating & 2482  & \multirow{2}{*}{53760} & \multirow{2}{*}{4080} &  \multirow{2}{*}{3640}\\
resting    & 2909  &&&\\
other      & 735   & 10616 & 1726   & 1575\\
\hline
total      & 13235 & 151478 & 11962 & 11262\\
\hline
\end{tabular}
\end{table}

\section{Model} \label{sec:model}

In this section, we provide an architectural overview of the proposed RNN classifiers, discuss the choice of the values for the hyper-parameters involved, and outline the related learning and evaluation processes as well as the utilized performance metrics.

\subsection{Architecture} \label{subsec:arch}

An RNN is an artificial neural network with its neurons organized into successive layers, i.e., input layer, hidden layer, and output layer. Each connection between neurons has a corresponding trainable weight. RNNs differ from feed-forward neural networks in that their hidden layer neurons are connected to themselves in a recurrent manner resembling a feedback loop. Therefore, the values of the weights associated with the hidden layer neurons at any time step $t$, i.e., the hidden state denoted by $\mathbf{h}_t$, are dependent on their values at the previous time step $t-1$, i.e., the previous hidden state denoted by $\mathbf{h}_{t-1}$, and the current input $\mathbf{x}_t$. This relationship can be expressed as
\begin{equation}
	\mathbf{h}_t = \mathbf{f}(\mathbf{x}_t, \mathbf{h}_{t-1}).
\end{equation}
The in-built recurrence enables RNNs to learn and recognize patterns over input sequences, extract meaningful features from sequential observations, and map the learned features to discriminate classes.

\begin{figure*}
	\centering
	\includegraphics[angle=90,scale=0.5]{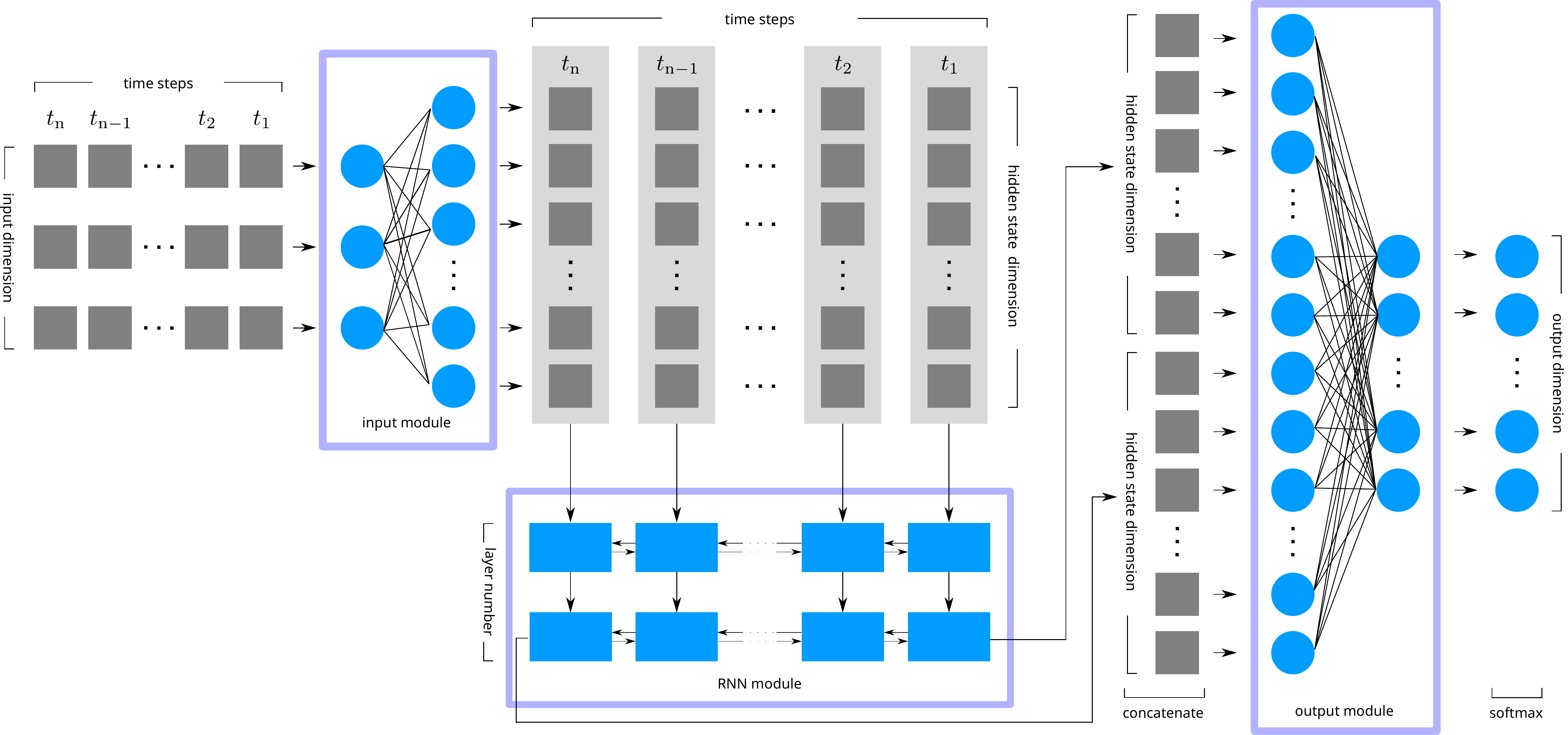}
	\caption{Model architecture of the considered RNN classifier. Blue circles represent the neurons of their respective perceptrons with ReLU activation functions and blue rectangles represent the RNN cells at each timestep.}
	\label{fig:rnn_architecture}
\end{figure*}

The base RNN classifier architecture that we use in this work is depicted in Fig.~\ref{fig:rnn_architecture}.
The input module is a perceptron that augments the dimension of the input data from three to the hidden-state dimension, which is a model hyper-parameter. The output of the input module is fed to the recurrent module, which learns a non-linear encoding of the input sequences. The output module takes the encoding produced by the recurrent module as input and shrinks its dimension down to the number of classes. If the recurrent module is bi-directional, two encoding vectors corresponding to the forward and backward directions are concatenated and fed into the output module. In the perceptrons of both input and output modules, the rectified linear unit (ReLU) activate function is used except for the last layer of the output module where the softmax activation function is used to predict the likelihood of each class.

\subsection{Hyper-parameters} \label{subsec:hype}

As the simple Elman/Jordan RNN often suffers from the vanishing/exploding gradient problem when dealing with long input sequences, gated RNN variants are more commonly used in practice. The first structural hyper-parameter of our RNN-based classifier is the gated RNN variant utilized, i.e., long short-term memory (LSTM) \citep{hochreiter1997long} or gated recurrent units (GRU) \citep{cho2014properties}. The LSTM network comprises a memory cell, an input gate, a forget gate, and an output gate. The memory cell accumulates the information extracted from the input sequences and the gates modulate the inward and outward flux of the information. The GRU network is similar to LSTM but does not have any output gate. Therefore, it has fewer parameters and is computationally less demanding compared to LSTM.

The bidirectional RNNs use an additional hidden layer to pass the information in the backward direction. This can enable more flexible processing of the information within the input sequence. The number of directions, i.e., the network being uni- or bi-directional is another structural hyper-parameter of our base RNN architecture. The other structural hyper-parameters are the number of stacked RNN layers and the number of neurons in each hidden layer. We consider the number of RNN layers to be either one or two and the number of hidden-layer neurons to be $32$, $64$, or $128$. We label the models that correspond to different combinations of the values considered for the structural hyper-parameters by hyphenating the number of directions (bi or uni), the RNN variant (LSTM or GRU), the number of layers ($1$ or $2$), and the number of hidden-layer neurons ($32$, $64$ or $128$) as shown in Table~\ref{tab:result}.

\subsection{Evaluation}

We compare the performance of the proposed models with that of two CNN-based time-series classification algorithms, which are proposed in~\citep{wang2017time} and are called the fully convolutional network (FCN) and the residual network (ResNet), in terms of both classification accuracy and model complexity. As shown in~\cite{fawaz2019deep}, these algorithms are among the best-performing existing time-series classification algorithms.

We evaluate the performance of all considered models using the four datasets explained in Section~\ref{sec:data} via a leave-one-animal-out cross-validation (CV) scheme. In each CV fold, we train each model using the data of all animals but one and evaluate the trained model using the data of the animal whose data was left out in training. We then aggregate the evaluation results of all folds to obtain our cross-validated accuracy results. We carry out a limited tuning of the algorithmic hyper-parameters of the proposed models, i.e., learning rate, weight decay, batch size, and number of training iterations, and use the same values for all models except the number of training iterations that is specific to each model. For FCN and ResNet, we use the hyper-parameter values prescribed in~\cite{fawaz2019deep}.

We use the Matthews correlation coefficient (MCC) as the metric for classification accuracy since it is suitable for unbalanced multi-class classification problems~\citep{gorodkin2004comparing}. MCC takes into account the true and false positives and negatives and is generally regarded as a balanced accuracy measure that can be used even if the classes are of very different sizes.

We quantify the complexity of each model using three metrics, namely, the number of parameters, the amount of memory occupied, and the number of multiplication operations required to perform inference on one data sample using the model. We calculate the memory usage of the model utilizing the tool provided at~\cite{jacobkimmel} and count the number of required multiplication operations by examining the tensor operations involved in each model.

The basic building blocks of all considered models are perceptron, LSTM, GRU, or convolution modules whose required multiplication counts are given by
\begin{align}
	M_{\rm perc.} & = D_{\rm in} \times D_{\rm out} \times L\\
	M_{\rm LSTM} & = (D_{\rm h}^2 \times 8 + D_{\rm h} \times 3) \times L\\
	M_{\rm GRU} & = (D_{\rm h}^2 \times 6 + D_{\rm h} \times 3) \times L\\
	M_{\rm conv.} & = K \times C_{\rm in} \times (L - K + 1) \times C_{\rm out}
\end{align}
where $L$ is the length of the input time-series, $D_{\rm in}$ and $D_{\rm out}$ are the dimensions of the input and output layers in the perceptron input/output modules, respectively, $D_{\rm h}$ is the dimension of the hidden layer in the RNN modules, $K$ is the size of the convolution kernel, and $C_{\rm in}$ and $C_{\rm out}$ are the numbers of the input and output channels in any convolutional layer, respectively.

We implement all models and algorithms using the PyTorch library for Python and run them on CSIRO's high-performance computing clusters.

\section{Results} \label{sec:eva}

We present the evaluation results of classification accuracy and model complexity for all considered models in this section.

\subsection{Accuracy}

\begin{table}[htbp] 
\caption{The MCC values of the proposed RNN-based and the state-of-the-art CNN-based classifiers evaluated using the considered datasets. The highest three MCC values for each dataset are shown in \textcolor{OliveGreen}{green}, \textcolor{RoyalBlue}{blue}, and \textcolor{BrickRed}{red}.} 
\label{tab:result}
\centering
\begin{tabular}{l c c c c}
\hline
& \multicolumn{4}{c}{dataset}\\
model & Arm18 & Lan19 & Arm20c & Arm20e \\
\hline
bi-LSTM-2-128  & 0.891 & \textcolor{OliveGreen}{0.826} & 0.880 & 0.789 \\
bi-LSTM-2-64   & 0.888 & 0.803 & 0.875 & 0.774 \\
bi-LSTM-2-32   & 0.778 & 0.755 & 0.857 & 0.751 \\
bi-LSTM-1-128  & 0.858 & 0.787 & 0.870 & 0.777 \\
bi-LSTM-1-64   & 0.772 & 0.751 & 0.864 & 0.760 \\
bi-LSTM-1-32   & 0.754 & 0.721 & 0.847 & 0.653 \\
uni-LSTM-2-128 & 0.879 & 0.806 & 0.806 & 0.789 \\
uni-LSTM-2-64  & 0.856 & 0.757 & 0.868 & 0.753 \\
uni-LSTM-2-32  & 0.829 & 0.746 & 0.852 & 0.446 \\
uni-LSTM-1-128 & 0.854 & 0.793 & 0.872 & 0.791 \\
uni-LSTM-1-64  & 0.777 & 0.741 & 0.861 & 0.739 \\
uni-LSTM-1-32  & 0.734 & 0.729 & 0.829 & 0.665 \\
\hline 
bi-GRU-2-128   & \textcolor{OliveGreen}{0.910} & \textcolor{RoyalBlue}{0.811} & \textcolor{OliveGreen}{0.895} & \textcolor{RoyalBlue}{0.800} \\
bi-GRU-2-64    & 0.891 & 0.800 & 0.884 & \textcolor{BrickRed}{0.797} \\
bi-GRU-2-32    & 0.883 & 0.753 & 0.876 & 0.759 \\
bi-GRU-1-128   & 0.886 & 0.789 & 0.884 & 0.779 \\
bi-GRU-1-64    & 0.857 & 0.766 & 0.871 & 0.776 \\
bi-GRU-1-32    & 0.757 & 0.738 & 0.851 & 0.747\\
uni-GRU-2-128  & \textcolor{RoyalBlue}{0.908} & \textcolor{BrickRed}{0.810} & \textcolor{RoyalBlue}{0.891} & 0.789 \\
uni-GRU-2-64   & 0.900 & 0.776 & 0.877 & 0.774 \\
uni-GRU-2-32   & 0.857 & 0.714 & 0.851 & 0.738 \\
uni-GRU-1-128  & 0.898 & 0.789 & \textcolor{BrickRed}{0.885} & 0.777 \\
uni-GRU-1-64   & 0.879 & 0.755 & 0.873 & 0.787 \\
uni-GRU-1-32   & 0.723 & 0.732 & 0.858 & 0.750 \\
\hline 
FCN            & 0.880 & 0.783 & 0.873 & 0.794 \\
ResNet         & \textcolor{BrickRed}{0.907} & 0.791 & 0.874 & \textcolor{OliveGreen}{0.809} \\
\hline
\end{tabular}
\end{table}

\begin{figure}[t]
	\centering
    \includegraphics[width=0.49\textwidth]{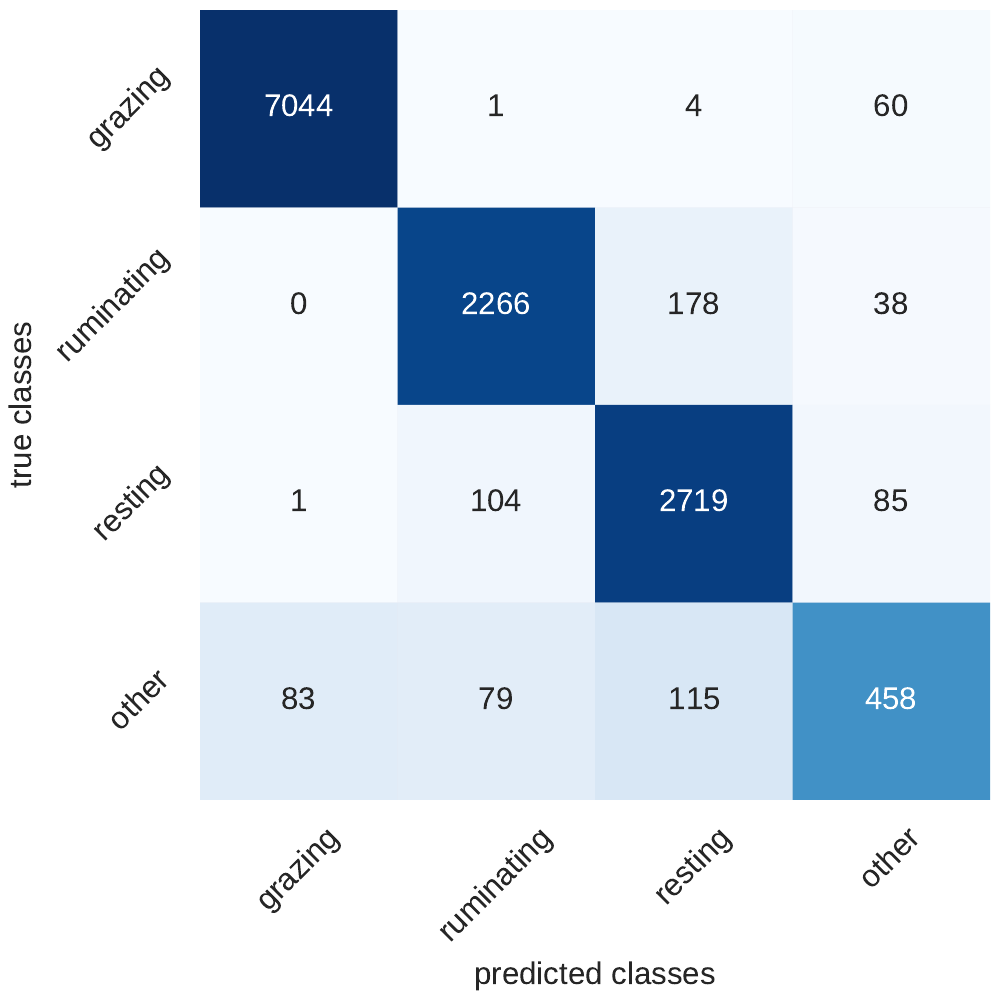}
    \includegraphics[width=0.49\textwidth]{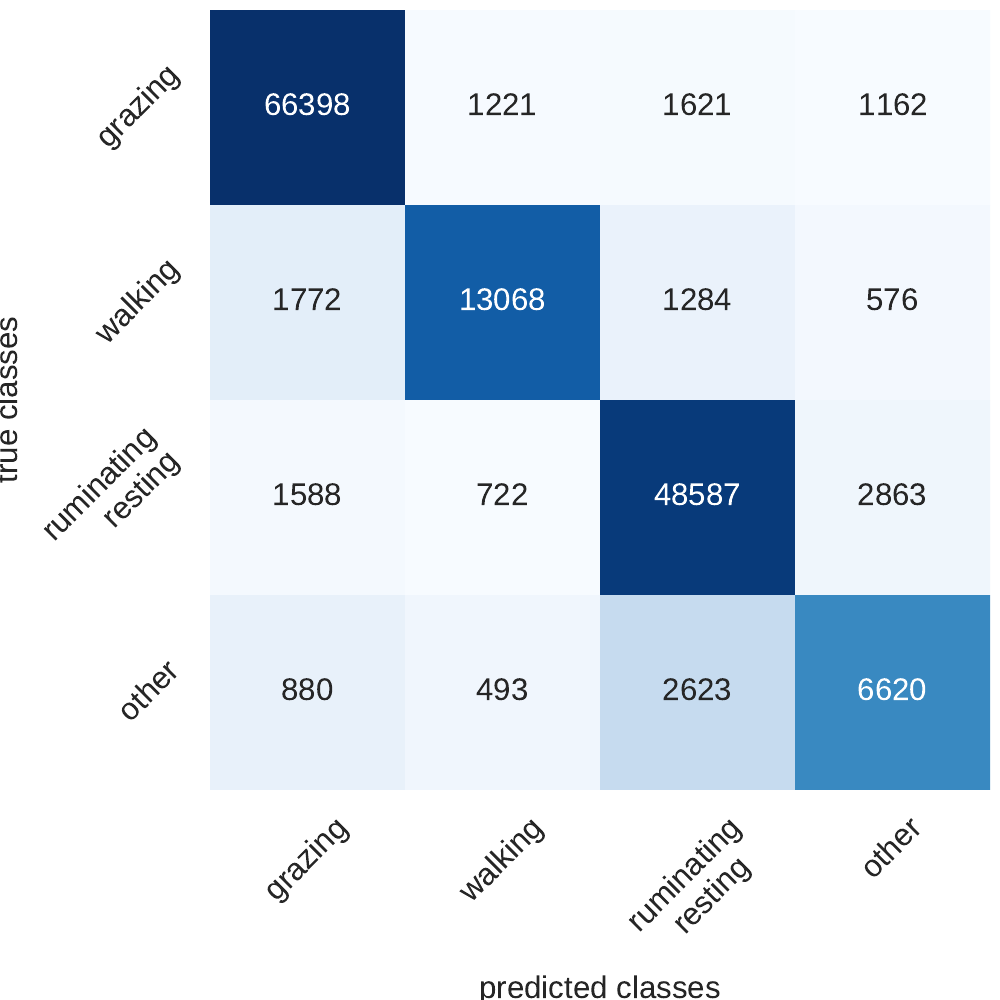}
    \includegraphics[width=0.49\textwidth]{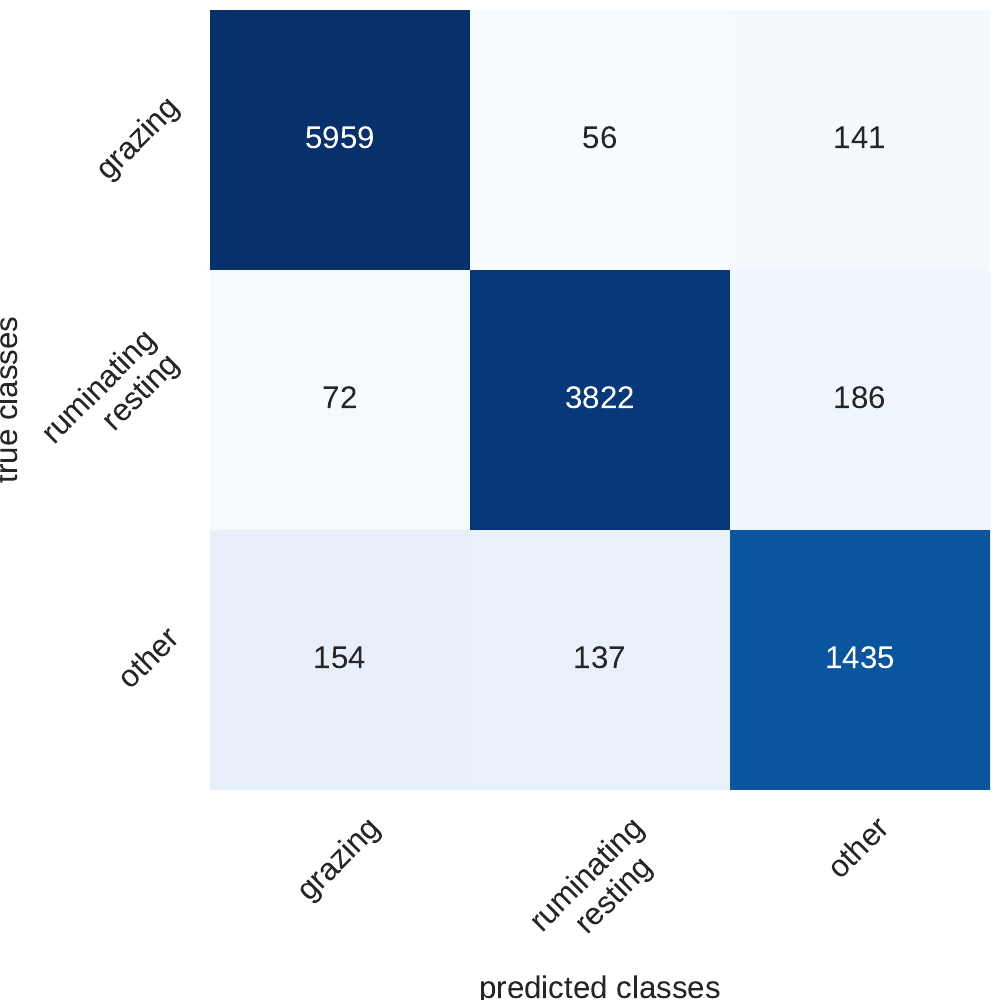}
    \includegraphics[width=0.49\textwidth]{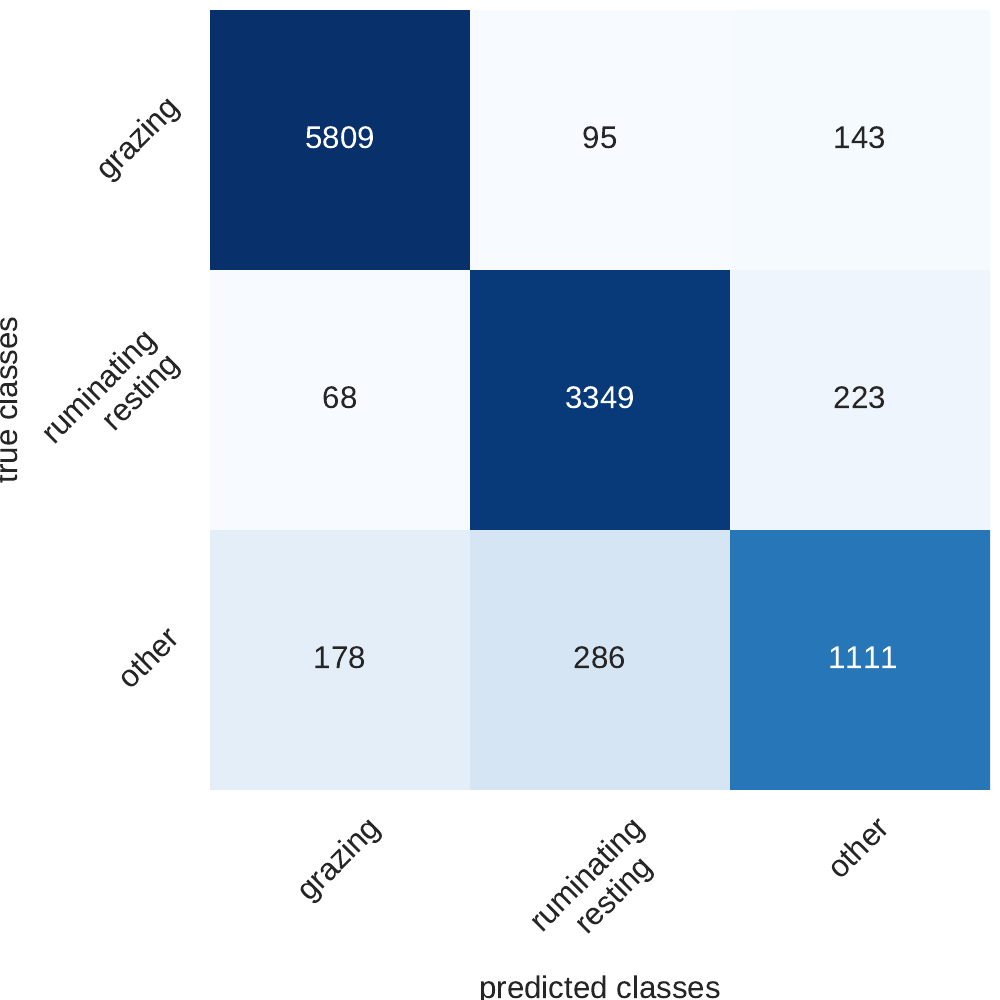}
	\caption[]{The confusion matrices corresponding to the best performing model for each dataset, top left: bi-GRU-2-128 for Arm18, top right: bi-LSTM-2-128 for Lan19, bottom left: bi-GRU-2-128 for Arm20c, bottom right: ResNet for Arm20e.}
	\label{fig:conf_mat_best}
\end{figure}

\begin{figure}[h]
	\centering
    \includegraphics[width=0.49\textwidth]{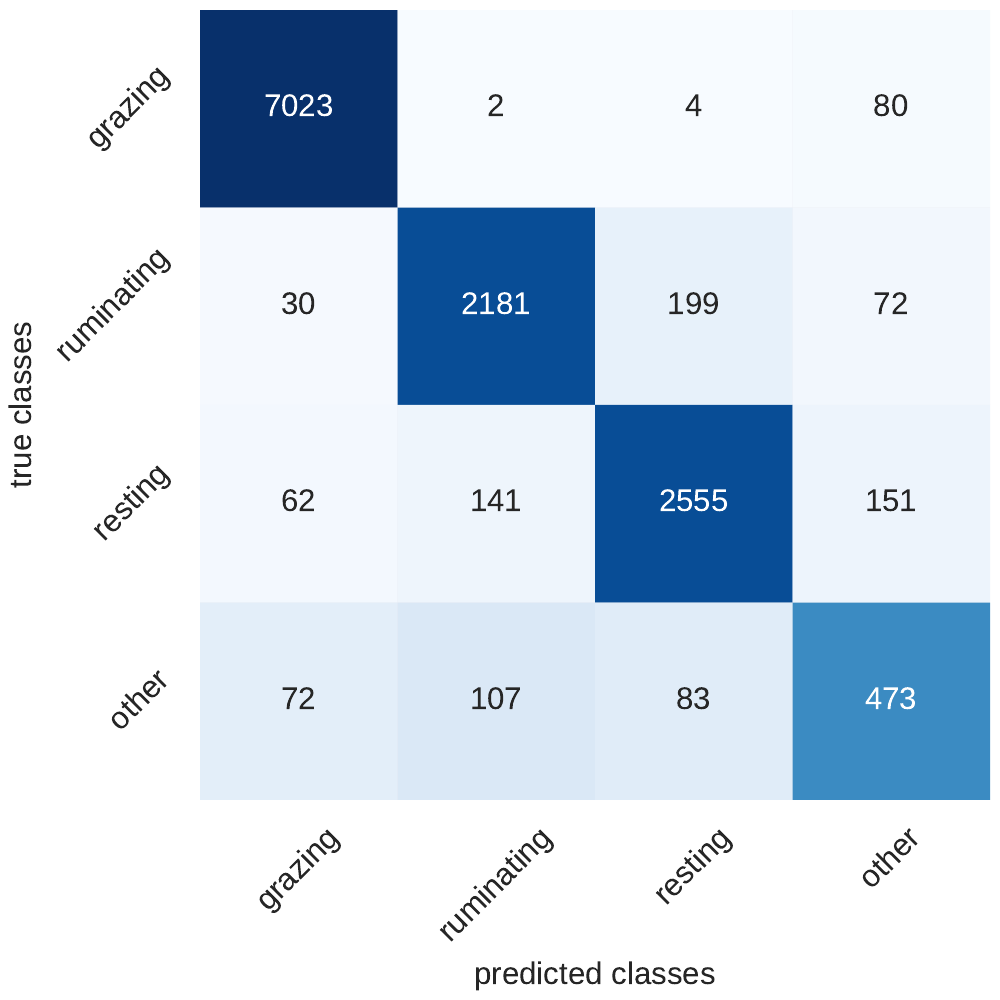}
    \includegraphics[width=0.49\textwidth]{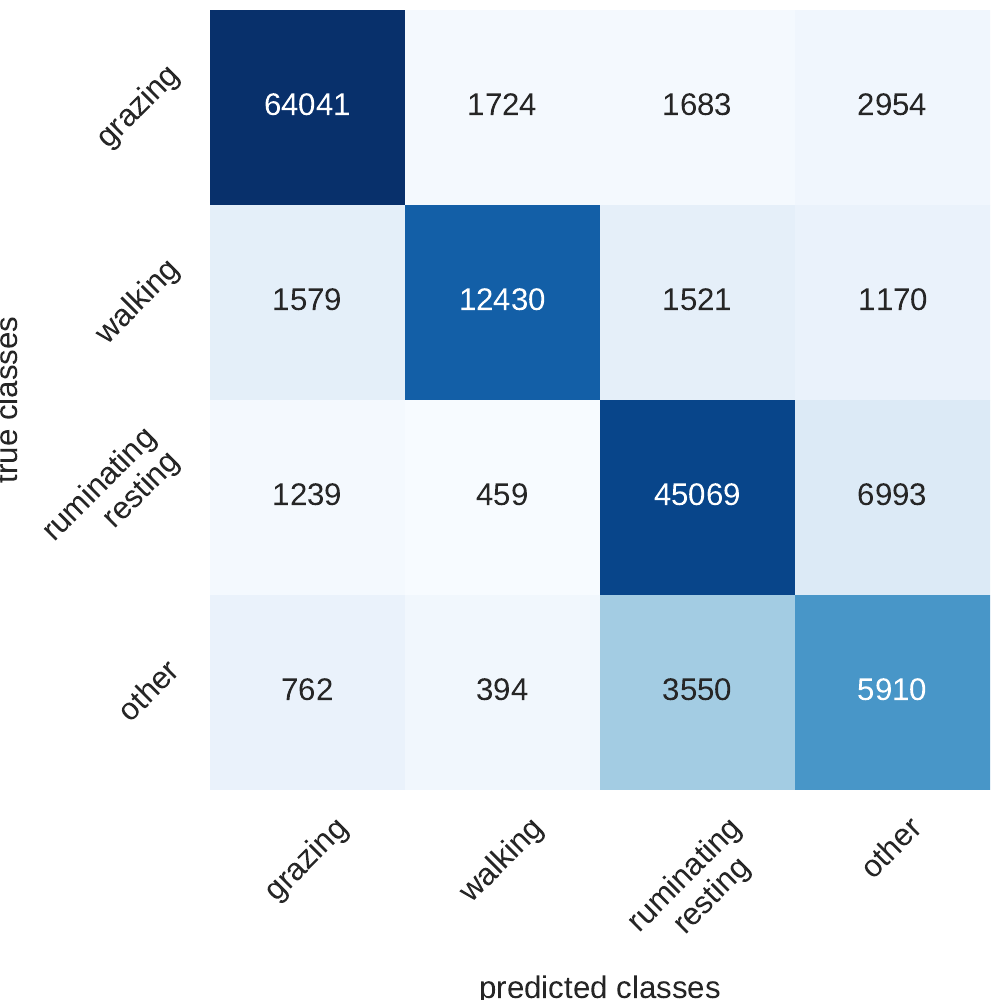}
    \includegraphics[width=0.49\textwidth]{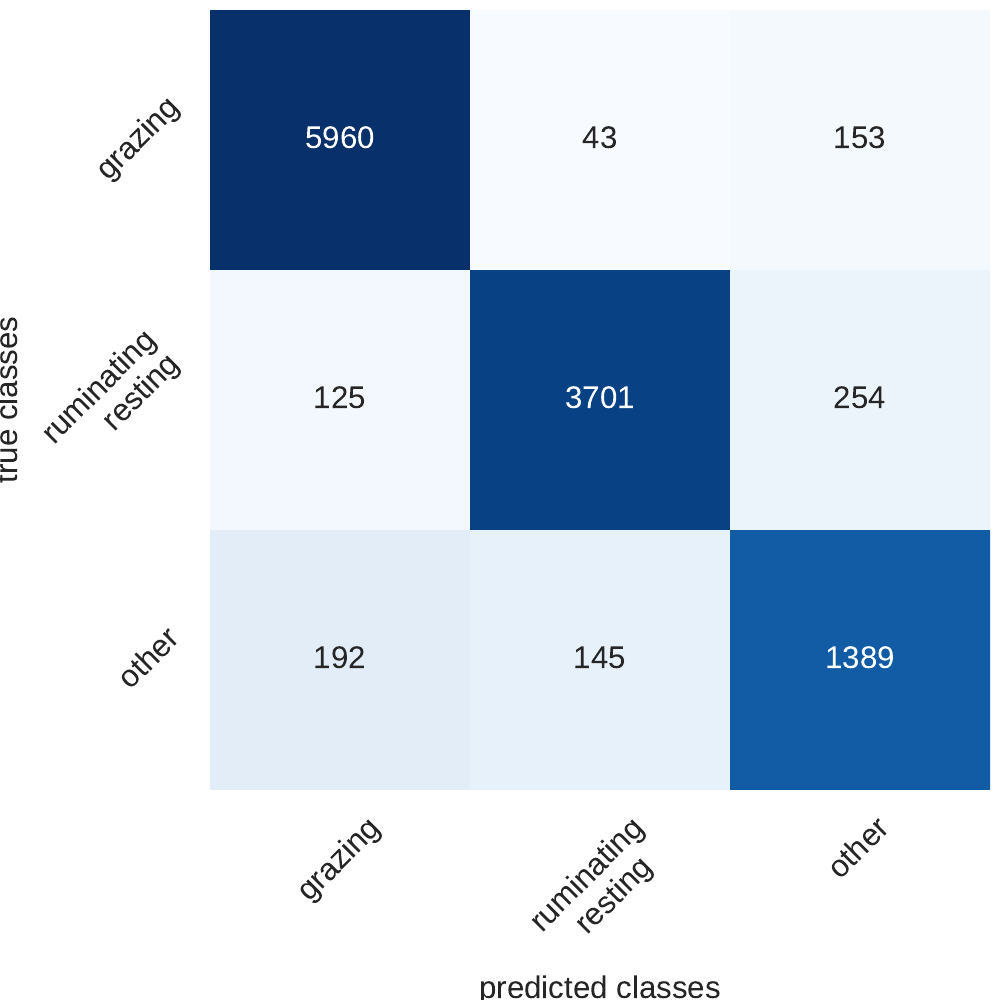}
    \includegraphics[width=0.49\textwidth]{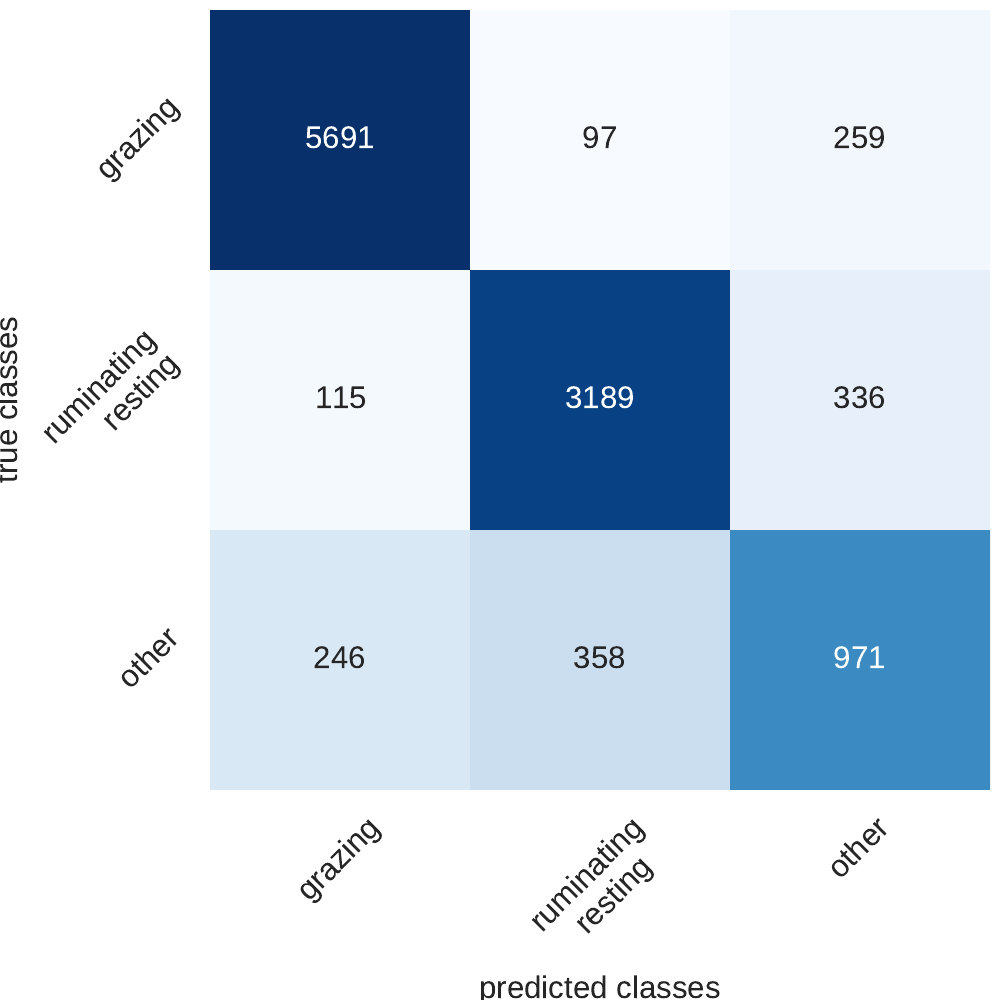}
	\caption[]{The confusion matrices corresponding to the uni-GRU-1-64 model for all datasets, top left: Arm18, top right: Lan19, bottom left: Arm20c, bottom right: Arm20e.}
	\label{fig:conf_mat_ug164}
\end{figure}

\begin{figure}
	\centering
	\includegraphics[scale=0.54]{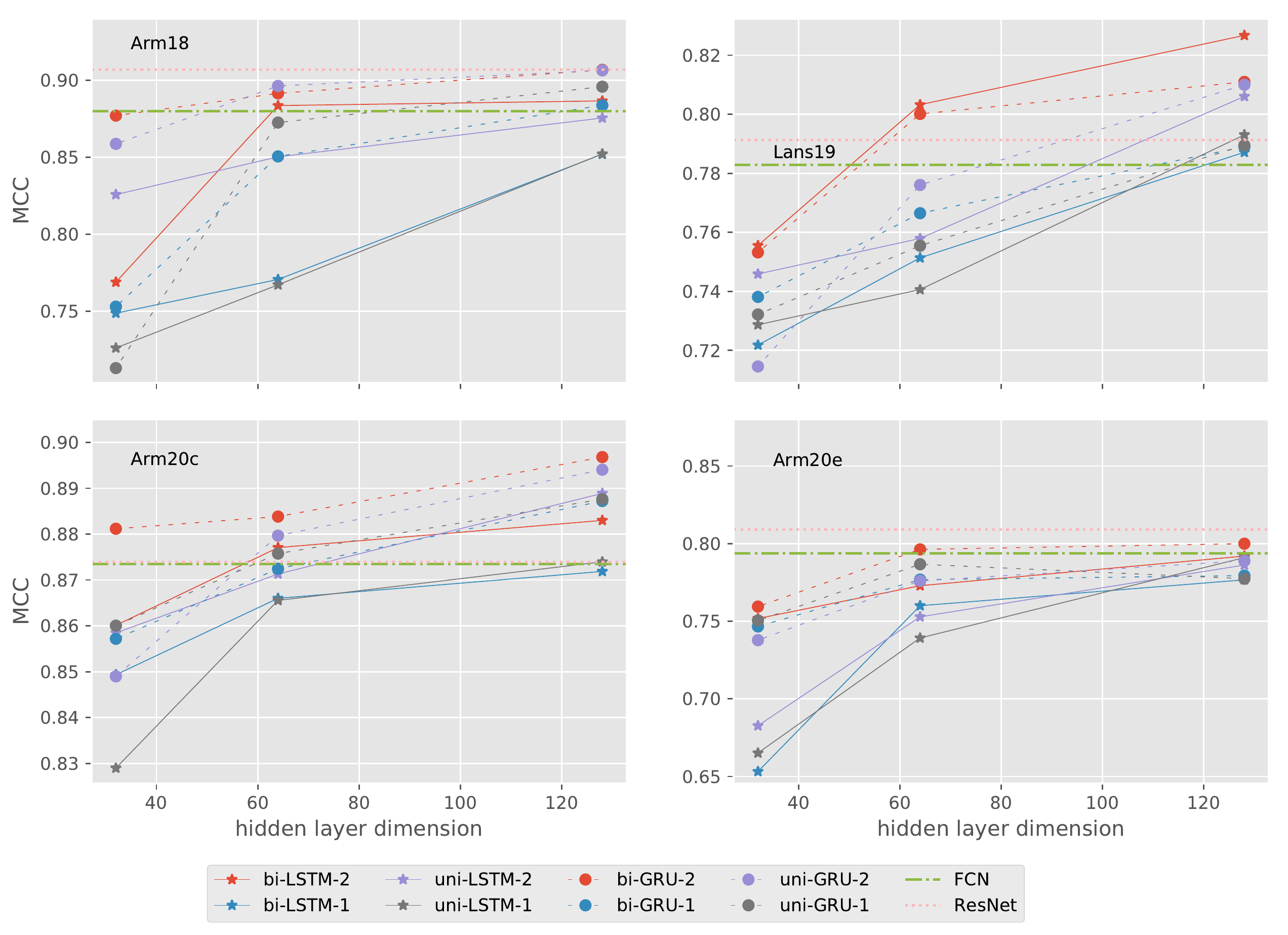}
	\caption{The MCC versus the hidden layer dimension for all considered RNN-based classification models and each dataset.}
	\label{fig:mcc_neuron}
\end{figure}

Table~\ref{tab:result} shows the MCC values representing the classification accuracy of the proposed RNN-based models and the considered state-of-the-art CNN-based models for each dataset. In Figs.~\ref{fig:conf_mat_best} and~\ref{fig:conf_mat_ug164}, for each considered dataset, we present the confusion matrices corresponding to the best performing model and the uni-GRU-1-64 model, respectively. We will explain the reason for choosing this model later in this section. To facilitate a visual comparison, we plot the values of the MCC against the dimension of the hidden layer for each dataset and all RNN-based models in Fig.~\ref{fig:mcc_neuron}.

The results show that the large RNN-based models, i.e., bi-LSTM-2-128 and bi-GRU-2-128, yield superior or similar performance to the CNN-based models. The GRU-based models are in general more accurate than the LSTM-based ones, although the difference is relatively small. The accuracy of the RNN-based models generally decreases as the the number of directions, layers, or neurons in each hidden layer decreases. However, the number of neurons appear to affect the accuracy more significantly compared with the number of directions or layers. Therefore, the MCC values of the uni-GRU-1-128 are rather close to those of the best-overall-performing model that is bi-GRU-2-128. The ResNet model is slightly more accurate than the FCN model. This is consistent with the results reported in~\cite{fawaz2019deep} for multivariate time-series classification.

\subsection{Complexity}

\begin{table}[htbp] 
\caption{The values of the model complexity metrics, i.e., the number of multiplication operations, the number of parameters, and the required memory, for the proposed RNN-based and the state-of-the-art CNN-based classifiers.}
\label{tab:complex}
\centering
\begin{tabular}{l c c c}
\hline 
model & operations (M) & parameters (K) & memory (MB)\\ 
\hline 
bi-LSTM-2-128 & 134.7 & 661   & 3.3\\ 
bi-LSTM-2-64  & 33.8  & 166.7 & 1.0 \\ 
bi-LSTM-2-32  & 8.5   & 42.3  & 0.4 \\ 
bi-LSTM-1-128 & 67.4  & 265.7 & 1.8 \\ 
bi-LSTM-1-64  & 16.9  & 67.3  & 0.7 \\ 
bi-LSTM-1-32  & 4.3   & 17.3  & 0.3 \\ 
uni-LSTM-2-128& 67.4  & 265.2 & 1.5 \\ 
uni-LSTM-2-64 & 16.9  & 67.1  & 0.5 \\ 
uni-LSTM-2-32 & 4.3   & 17.2  & 0.2 \\ 
uni-LSTM-1-128& 33.8  & 133.1 & 1.0 \\ 
uni-LSTM-1-64 & 8.5   & 33.8  & 0.4 \\ 
uni-LSTM-1-32 & 2.1   & 8.7   & 0.2 \\ 
\hline 
bi-GRU-2-128  & 101.2 & 496.1 & 2.7 \\ 
bi-GRU-2-64   & 25.4  & 125.2 & 0.9 \\ 
bi-GRU-2-32   & 6.4   & 31.9  & 0.3 \\ 
bi-GRU-1-128  & 50.6  & 199.7 & 1.5 \\ 
bi-GRU-1-64   & 12.7  & 50.7  & 0.6 \\ 
bi-GRU-1-32   & 3.2   & 13.1  & 0.3 \\ 
uni-GRU-2-128 & 50.7  & 199.2 & 1.3 \\ 
uni-GRU-2-64  & 12.7  & 50.4  & 0.5 \\ 
uni-GRU-2-32  & 3.2   & 12.9  & 0.2 \\ 
uni-GRU-1-128 & 25.4  & 100.1 & 0.9 \\ 
\textcolor{OliveGreen}{uni-GRU-1-64}  & 6.4   & 25.5  & 0.4 \\ 
uni-GRU-1-32  & 1.6   & 6.6   & 0.2 \\ 
\hline 
FCN           & 67.9  & 267.3 & 3.0 \\ 
ResNet        & 132.6 & 520.2 & 4.6 \\ 
\hline 
\end{tabular}
\end{table}

\begin{figure}
	\centering
	\includegraphics[scale=0.53]{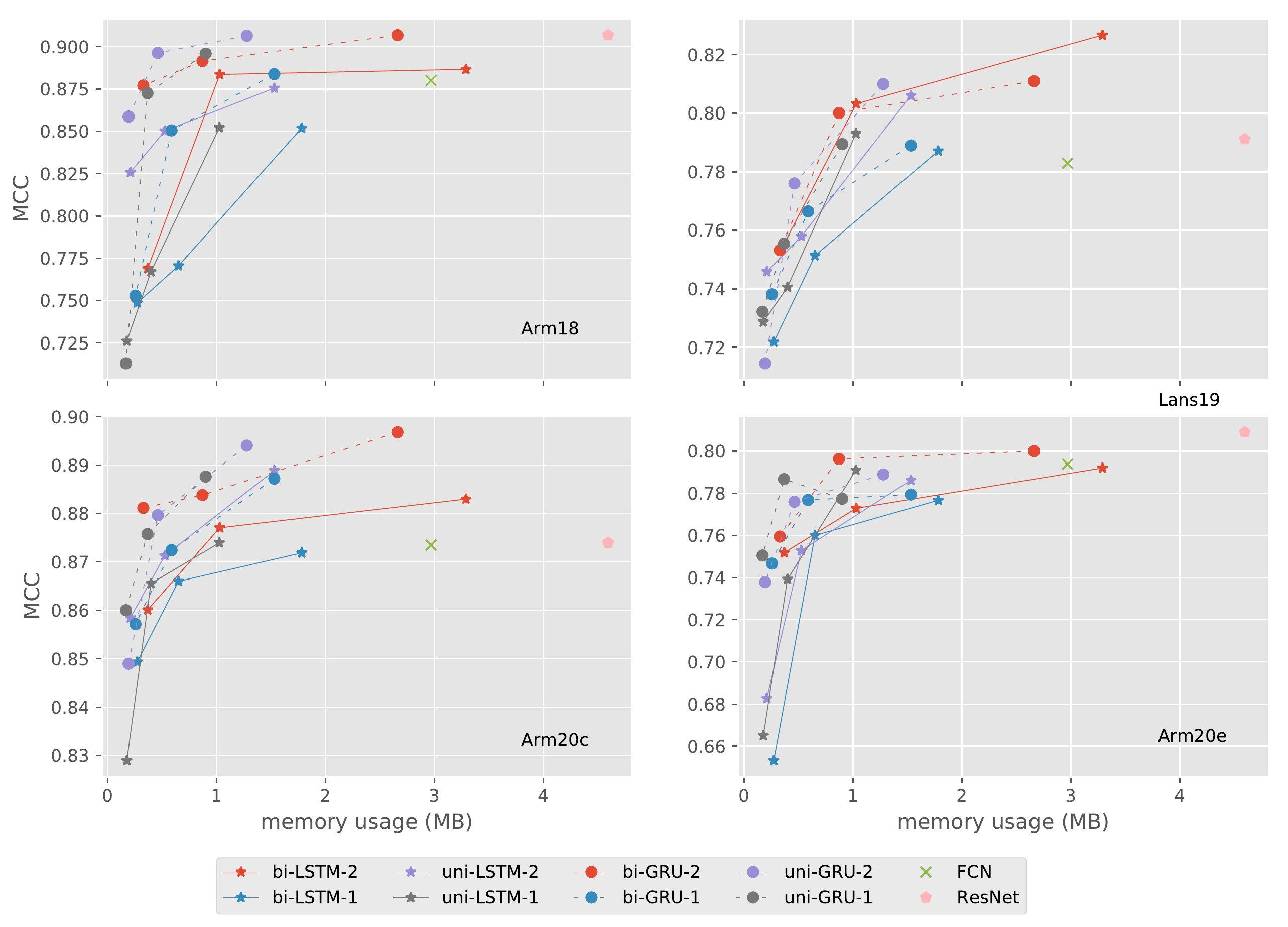}
	\caption{The MCC versus the memory usage for all considered classification models and datasets.}
	\label{fig:mcc_memory}
\end{figure}

\begin{figure}
	\centering
	\includegraphics[scale=0.53]{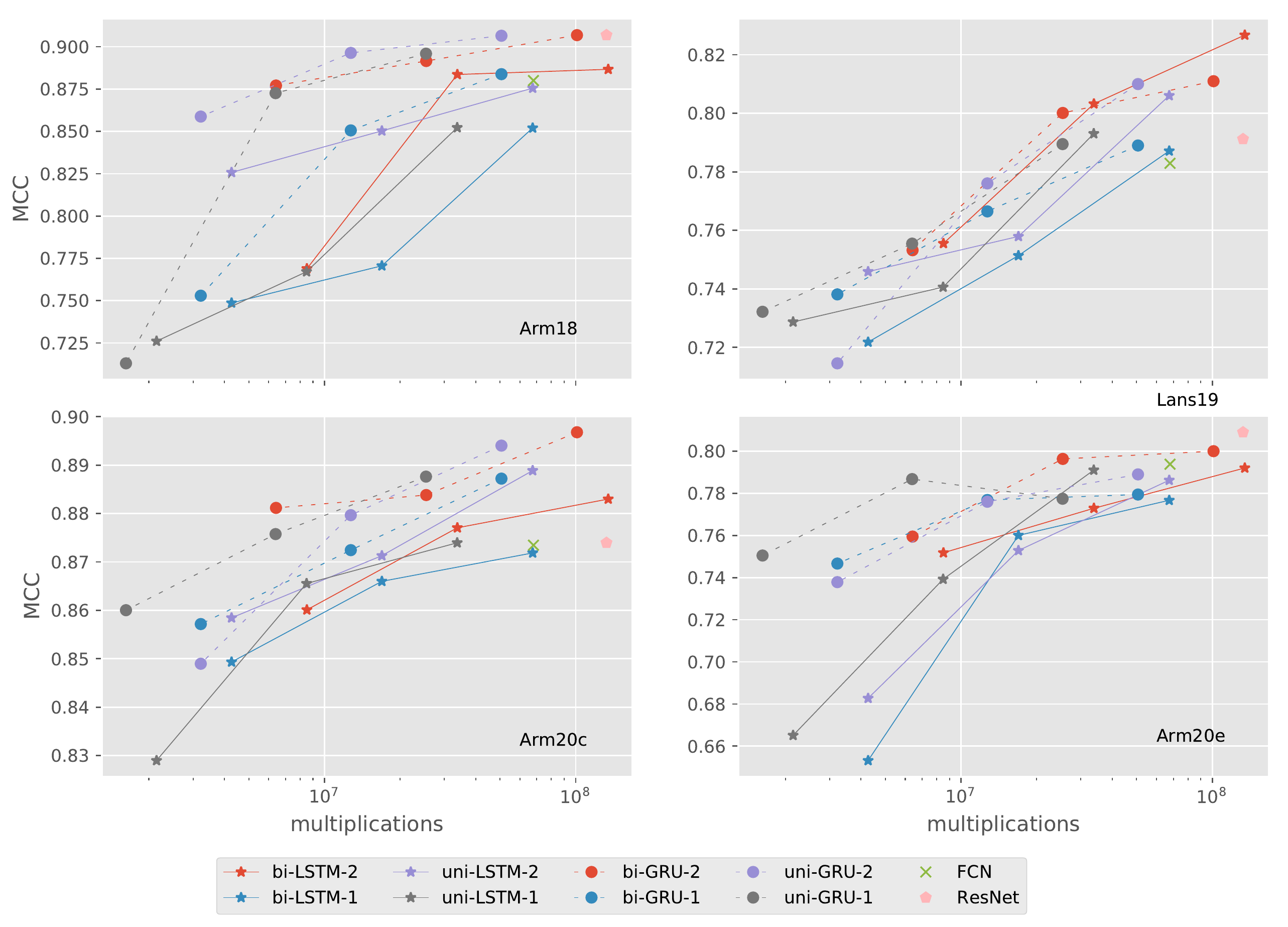}
	\caption{The MCC versus the number of multiplication operations required for inference (a single forward pass) for all considered classification models and datasets.}
	\label{fig:mcc_multi}
\end{figure}

In Table~\ref{tab:complex}, we present the values of the model complexity metrics, i.e., the number of multiplication operations required to perform inference on a single input data sample, the number of trainable parameters, and the memory usage, for the proposed RNN-based and the considered state-of-the-art CNN-based classifiers. We calculate these values considering the number of classes to be four. To provide a visual representation of the accuracy-complexity trade-off underlying the examined models, we plot the values of MCC against the memory usage and the number of required multiplications for all models and datasets in Figs.~\ref{fig:mcc_memory} and~\ref{fig:mcc_multi}, respectively. In these figures, the points with the same shape and color correspond to the models that only vary in the hidden layer dimension. The larger the dimension, the higher the memory usage and multiplication count. Note that the complexity of each model for performing inference is independent of the dataset used as long as the data samples have the same size, which is the case here.

Among all models, bi-LSTM-2-128 is the most complex one in terms of the number of parameters and multiplication operations. Its memory usage is also the highest among the RNN-based models. The bi-GRU-2-128 model is almost three-fourths as complex as the bi-LSTM-2-128 model in terms of all complexity metrics despite having better accuracy. Unidirectional models are significantly less complex compared to their bidirectional counterparts as their complexity metric values are less than half of those of the corresponding bidirectional models. However, this does not come at the expense of any significant loss of accuracy. We observe a similar trend when examining the models with one hidden layer versus the corresponding models with two hidden layers. 

The complexity difference in models that only differ in the number of hidden layer neurons (hidden layer dimension) is more pronounced as going from $128$ to $64$ reduces the complexity metrics by almost four folds. In general, the models with $64$ neurons in their hidden layers are only slightly less accurate compared to their corresponding ones with $128$ neurons. However, reducing the hidden layer dimension to $32$ results in a more noticeable decrease in accuracy. Therefore, the uni-GRU-1-64 model appears to strike the best balance between accuracy and performance. When contrasted to the state-of-the-art CNN-based models, FCN and ResNet, the uni-GRU-1-64 has a comparable accuracy but is substantially less complex as the values of its complexity metrics are more than an order of magnitude smaller than those of the FCN and ResNet models.

\section{Discussion} \label{sec:discuss}

A main feature of the classification models that we have studied in this work is their ability to be trained and perform inference in an end-to-end manner. Specifically, they take the multivariate time-series data of triaxial accelerometer readings as the input and produce the predicted animal behavior class as the output. The conventional models commonly adopted prior to the introduction of the DNN-based end-to-end models require scrupulous feature engineering in conjunction with expert domain knowledge. The input to the classification model in such approaches is the extracted features and not the raw data. The raw data is fed to a preceding processing stage called feature extraction and generates the features. This approach involve two rather separate stages. The first stage is to determine the best features to extract from the data and the second stage is to find the best classification model that maps the extracted features to the desirable output.

In the RNN- and CNN-based models studied in this work, the RNN modules or the CNN layers can be viewed to play the role of feature extraction. However, the parameters (weights) of the RNN/CNN layers are trainable. This mean the features are learned jointly with the discriminator (classifier) in an efficient way. This is the key to their superior performance and wide acceptance in recent years to tackle many problems in various applications.

Our proposed RNN-based models exhibit favorable performance when evaluated using the considered performance metrics and contrasted with the state-of-the-art CNN-based models. However, they suffer from a fundamental limitation that is inherent to all RNN-based models. The recurrent operations of an RNN are performed sequentially and may not be executed in parallel. On the other hand, feed-forward neural networks including the CNNs predominantly involve arithmetic operations over tensors, which can be performed in parallel given the suitable compute capability. Therefore, RNN-based models can generally benefit less from parallelization compared with the CNN-based models. Nonetheless, parallel processors that can accelerate tensor operations are resource-hungry and still rarely integrated into edge devices or embedded systems. This means the sequential nature of the RNNs does not pose any substantial disadvantage to our proposed RNN-based models when the target platform is an edge or embedded device.

To evaluate the performance of our proposed models, we used four datasets, three of which are collected using collar tags and one using ear tags. The datasets are also collected at two different research facilities, i.e., three datasets at one facility and one dataset at another. The locations of the facilities are more than 1,300km apart. This has resulted in a significant diversity among the datasets for the following reasons. First, the utilized collar and ear tags are placed/attached to different parts of cattle's body and their accelerometers are subject to significantly different acceleration patterns when animals exhibit various considered behaviors. The type of the accelerometers used and their sampling rates are also different in the collar and ear tags. In addition, the impact of earth's gravity on the accelerometer readings of the two devices is different as the orientation of the collar and ear tags correlate with each considered behavior in distinct ways. Second, the dimensions of the area in which the experiments took place as well as the climate, pasture type and quality, and terrain characteristics are substantially different in the two utilized research facilities. We have also used several different breeds of cattle in the data collection experiments. The important observation here is that the proposed models perform well with all four datasets despite their dissimilarities. Therefore, the proposed algorithms are robust to a broad range of variations in the data due to how, where, and when it is sensed.

\section{Conclusion} \label{sec:conclusion}

We proposed a set of end-to-end RNN-based algorithms for animal behavior classification using triaxial accelerometry data. We examined the performance of the proposed models in terms of classification accuracy and model complexity in comparison with two state-of-the-art CNN-based models recently proposed for classifying multivariate time-series data. In our performance evaluations, we used four datasets collected from grazing cattle using collar or ear tags containing triaxial accelerometer sensors. We also utilized a leave-one-animal-out cross-validation scheme. We considered the two most popular variants of RNN, namely, LSTM and GRU. We found that the GRU-based models achieve better classification accuracy compared to the LSTM-based models, with all considered datasets, despite their lower complexity. We also observed that the unidirectional models with a single hidden layer are only slightly less accurate than their bidirectional or double-hidden-layer counterparts. Both accuracy and complexity were more sensitive to the dimension of the hidden layers. Overall, the unidirectional GRU-based model with one hidden layer of size $64$ appeared to offer the best balance between accuracy and complexity. This RNN-based model is nearly as accurate as the state-of-the-art CNN-based FCN and ResNet models but with substantially reduced computational and memory complexity. This makes it comparatively more suitable for implementation on resource-constrained edge devices or embedded systems. In future work, we will explore implementing the considered models on the embedded systems of our sensor devices, i.e., the collar and ear tags, and examine their on-device performance and the associated trade-offs.

\section*{Acknowledgement} \label{sec:acknowledgement}

This research was undertaken with strategic investment funding from the CSIRO and NSW Department of Primary Industries. We would like to thank the following technical staff who were involved in the research at CSIRO FD McMaster Laboratory Chiswick: Alistair Donaldson and Reg Woodgate with NSW Department of Primary Industries, and Jody McNally and Troy Kalinowski with CSIRO Agriculture and Food. In addition, we acknowledge the CSIRO staff who have contributed to the research projects at Lansdown Research Station that have produced one of the datasets used in this paper, especially, Mel Matthews, Holly Reid, Wayne Flintham, and Steve Austin. We also recognize the contributions of the CSIRO Data61 staff who have designed and built the hardware and software of the devices used for data collection, specifically, Lachlan Currie, John Scolaro, Jordan Yates, Leslie Overs, and Stephen Brosnan.

\bibliographystyle{elsarticle-num-names} 

\bibliography{lstm_classifier}

\end{document}